\documentclass[runningheads]{llncs}

\usepackage[ID=25]{eccv}

\usepackage{eccvabbrv}

\usepackage{graphicx}
\graphicspath{{figures/}}
\usepackage{booktabs}
\usepackage{amssymb}   %
\usepackage{wrapfig}   %
\usepackage{xcolor}
\usepackage{array}     %
\usepackage{tikz}      %
\usepackage{afterpage}
\usetikzlibrary{arrows.meta, calc, positioning, shapes.geometric, fit, backgrounds}

\ifdefined\pdfximage
  \usepackage[accsupp]{axessibility}  %
\fi

\usepackage{hyperref}

\usepackage{orcidlink}

\begin{document}

\title{Integrating Multi-view Multi-light Surface Reconstruction into Cultural Heritage Workflows}

\titlerunning{Multi-view Photometric Stereo for Cultural Heritage}

\author{Baptiste Brument\inst{1}\orcidlink{0009-0004-5372-2215} \and
Robin Bruneau\inst{2}\orcidlink{0000-0003-3178-9019} \and
Benjamin Coupry\inst{1}\orcidlink{0009-0008-7436-6888} \and
Vincent Demoulin\inst{3} \and
Jean M\'elou\inst{4}\orcidlink{0009-0000-7990-2290} \and
Antoine Laurent\inst{5}\orcidlink{0000-0002-3727-5644} \and
Fabien Castan\inst{3}\orcidlink{0009-0008-2282-7128} \and
Jean-Denis Durou\inst{1}\orcidlink{0009-0006-2626-7434} \and
Lilian Calvet\inst{6}\orcidlink{0000-0002-2565-2297}}

\authorrunning{B. Brument et al.}

\institute{IRIT, UMR 5505, CNRS, Toulouse, France \and
University of Zurich, Zurich, Switzerland \and
The Mill, France \and
Fittingbox, Lab\`ege, France \and
TRACES, UMR 5608, CNRS, Toulouse, France \and
ROCS, Balgrist University Hospital, University of Zurich, Zurich, Switzerland}

\maketitle

\begin{abstract}
  Cultural heritage documentation increasingly relies on image-based 3D surface reconstruction, with photogrammetry software making such workflows accessible to archaeologists, conservators, and heritage technicians. These tools have been successful for conventional multi-view acquisition, but they do not routinely exploit richer multi-view, multi-light data, despite its potential for improving fine-scale surface reconstruction. This limitation is particularly relevant in heritage contexts, where controlled-light acquisition devices such as RTI domes are already used to capture illumination-varying image sets. The challenge is therefore to connect these existing acquisition practices with recent computer vision methods in a form that can be used within operational heritage workflows.
  In this work, we address this need by integrating state-of-the-art components from computer vision for multi-view, multi-light surface reconstruction into Meshroom, an open-source photogrammetry framework. Rather than proposing a new reconstruction algorithm, our contribution is to assemble and expose existing advanced methods, namely a complete photometric stereo ecosystem (calibrated, self-calibrated and universal), automatic object masking, and multi-view normal-and-reflectance integration, within a usable heritage-oriented workflow. The proposed system thus provides an intermediate software layer between computer vision research code and practical cultural heritage applications, making recent techniques easier to use and evaluate.
  \keywords{Multi-view Photometric Stereo \and Photometric Stereo \and Cultural Heritage \and Photogrammetry}
\end{abstract}

\section{Introduction}
\label{sec:intro}

High-fidelity 3D replicas have become important tools across cultural heritage practice, and advances in sensors, acquisition, and reconstruction algorithms now produce increasingly detailed digital representations of heritage objects.
This precision matters because, for many artifacts, the relevant information lies in the fine surface relief, namely engravings, tool marks, and micro-relief, as much as in the overall shape. Such objects are also among the hardest to digitise, since they are frequently poorly textured, as with lithic, bone or uniform ceramics.

\begin{figure}[t]
  \centering
  \includegraphics[width=\textwidth]{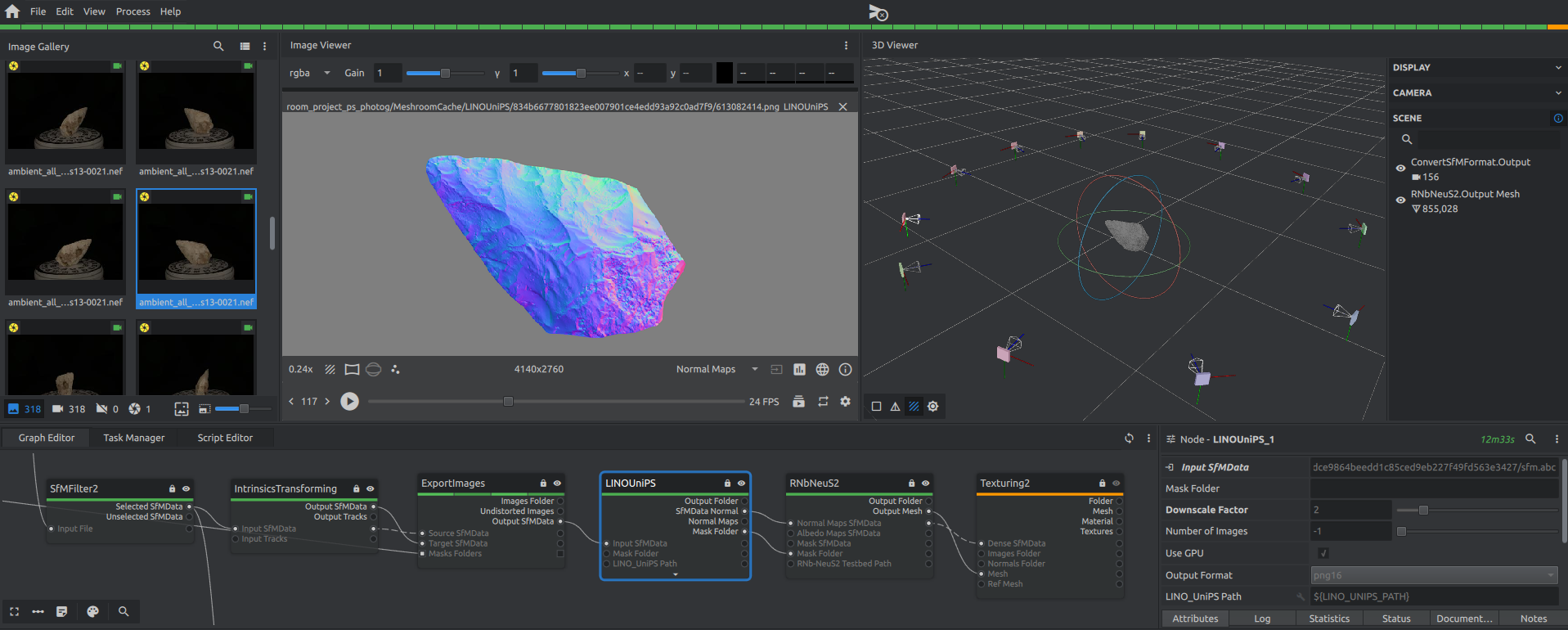}\\[3pt]
  \includegraphics[width=\textwidth]{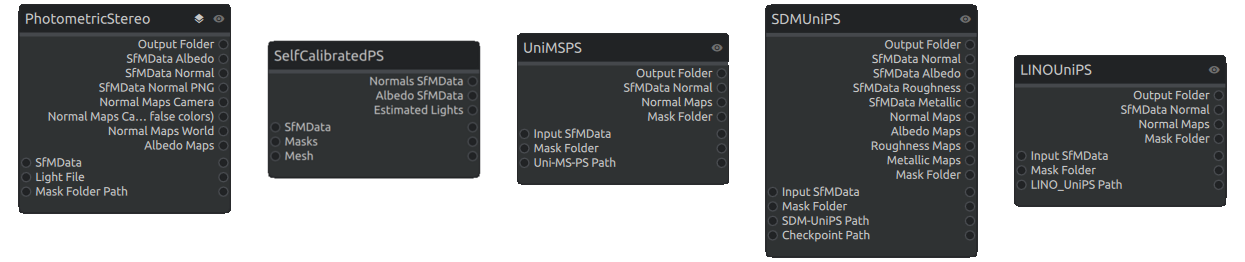}
  \caption{Our contribution: integrating a user-friendly multi-view, multi-light surface reconstruction pipeline into Meshroom~\cite{meshroom} for cultural heritage practitioners. \emph{Top}: reconstructing a lithic biface; the multi-view, multi-light inputs in the gallery (left), a per-view normal map from photometric stereo in the image viewer (centre), the final surface in the 3D viewer (right), and the processing graph (bottom; see \cref{fig:overview}). \emph{Bottom}: the photometric stereo methods integrated as interchangeable Meshroom nodes~\cite{mrunimsps,mrsdmunips,mrlinounips}.
  }
  \label{fig:teaser}
  \vspace{-10pt}
\end{figure}

To record this surface from photographs, the heritage community relies on two complementary families. Photogrammetry recovers 3D geometry from many viewpoints through mature, widely used software~\cite{comte2024notredame,realityscan,metashape}, but it is most reliable on global shape, struggles with uniform and non-Lambertian surfaces~\cite{karami2021noncollaborative}, and bakes the capture-time illumination into its texture (\cref{fig:shading}).
Reflectance transformation imaging (RTI), by contrast, photographs a fixed viewpoint under many light directions~\cite{malzbender2001ptm,mudge2006rti,earl2010rti}: varying the light reveals fine relief that flat photographs miss, and is widely used to read faint engravings and tool marks~\cite{pintus2019mlic,mansouri2026rti}, yet it yields an image-space, single-viewpoint representation, not a 3D surface.

These two families capture complementary cues, multi-view geometry and illumination-varying appearance, yet they are mostly acquired and analysed separately, and combining them remains a recognised open challenge in the heritage community~\cite{pamart2019rti,pamart2020taco}. The computer vision community has long sought to reconcile them in a single representation: multi-view, multi-light surface reconstruction, known as multi-view photometric stereo (MVPS), recovers per-view normal and reflectance maps from the multi-light images of each viewpoint (photometric stereo) and fuses them across views into one 3D surface (multi-view normal integration). Its key practical appeal is to reuse the illumination-varying images the heritage community already captures for RTI, now from several viewpoints, building on familiar acquisition practice rather than a new modality.
However, this reconstruction long proved difficult: early methods had to reconcile potentially conflicting depth and normal objectives and tended to erase the very detail they were meant to preserve~\cite{KayaKOFG22,KayaKOFG23}. Recent neural surface reconstruction removes this conflict through a single optimisation objective~\cite{rnbneus,rnbneus2,cao2024supernormal}, while photometric stereo itself now handles non-Lambertian materials and recovers per-pixel normals without prior light calibration~\cite{unimsps,sdmunips,linounips}. What remains is that these advances live as research prototypes and unpackaged code, out of reach of most heritage practitioners.

This lack of integration is itself the barrier to adoption, in a community that has nonetheless already embraced photogrammetric software.
We address it by integrating state-of-the-art computer vision components for multi-view, multi-light surface reconstruction into Meshroom~\cite{meshroom}, an existing open-source photogrammetry software, so that classical and recent photometric stereo and multi-view normal integration methods become usable and evaluable by heritage practitioners.
Moving these methods out of the laboratory means supporting many acquisition setups rather than a single one, as the constraints vary widely (access to the artefacts, availability and cost of a calibrated dome, indoor versus outdoor capture); we therefore expose a modular ecosystem of interchangeable nodes that adapts to each setup, from a calibrated dome to a hand-held flash in the field.
Concretely, our contributions are:
\begin{itemize}
  \item \textbf{(C1)} A complete photometric stereo ecosystem exposed as Meshroom nodes, covering calibrated photometric stereo with automatic calibration-sphere detection and light calibration, self-calibrated near-light photometric stereo~\cite{coupryssvm}, and universal neural methods that need no prior light calibration, namely UniMS-PS~\cite{unimsps}, SDM-UniPS~\cite{sdmunips} and LINO-UniPS~\cite{linounips};
  \item \textbf{(C2)} Automatic object masking~\cite{carion2025sam3segmentconcepts,mrsegmentation} and a multi-view normal integration node that optionally fuses the estimated reflectance, provided both as the reference RNb-NeuS2~\cite{rnbneus2} and as our permissively licensed OpenRNb~\cite{openrnb} reimplementation, which together close the chain from multi-light images to a single textured, high-fidelity surface mesh;
  \item \textbf{(C3)} An open-source release of all the nodes and of ready-to-use multi-view, multi-light pipelines, with sample code that promotes the integration of further photometric stereo and multi-view normal integration methods;
  \item \textbf{(C4)} A qualitative validation on real archaeological objects.
\end{itemize}

\vspace{5em}
\section{Related Work}
\label{sec:related}

\begin{wrapfigure}[23]{R}{0.34\textwidth}
  \vspace{-28pt}
  \centering
  \setlength{\tabcolsep}{1pt}%
  \renewcommand{\arraystretch}{0.9}%
  \begin{tabular}{@{}>{\centering\arraybackslash}m{0.05\linewidth} *{2}{>{\centering\arraybackslash}m{0.45\linewidth}}@{}}
    & \scriptsize Photograph & \scriptsize Reconstruction \\[1pt]
    \rotatebox{90}{\scriptsize Front} &
      \includegraphics[width=\linewidth]{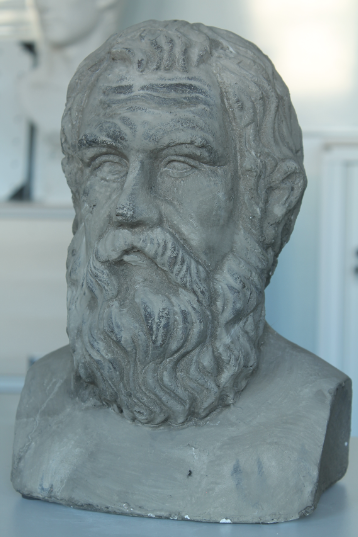} &
      \includegraphics[width=\linewidth]{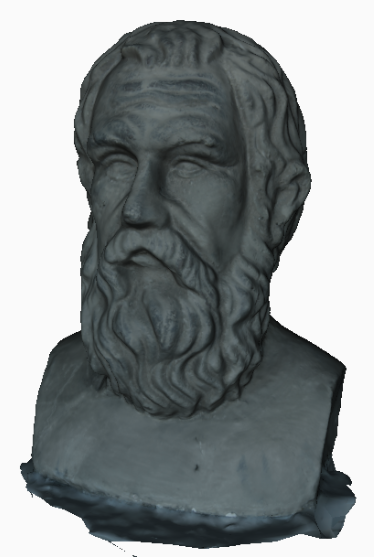} \\[2pt]
    \rotatebox{90}{\scriptsize Back} &
      \includegraphics[width=\linewidth]{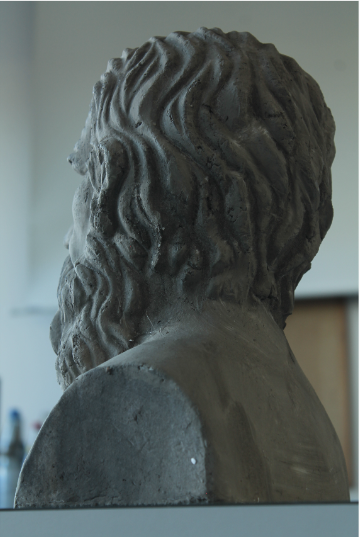} &
      \reflectbox{\includegraphics[width=\linewidth]{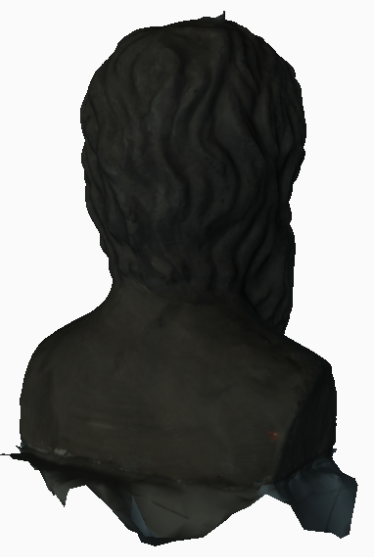}} \\
  \end{tabular}
  \caption{Texture from Meshroom's standard photogrammetry pipeline bakes in the acquisition lighting: lit more at the front, this Socrates bust reconstructs darker at the back despite an identical material.}
  \label{fig:shading}
\end{wrapfigure}

\paragraph{Photogrammetry for cultural heritage.}
Image-based 3D reconstruction has become the dominant photographic technique in heritage documentation~\cite{remondino2006imagebased,remondino2011heritage}, largely because computer vision methods have been translated into robust, easy-to-use software that fits heritage workflows, from single artefacts to large campaigns such as archaeological excavations~\cite{dereu2014introducing}, monument documentation~\cite{yastikli2007documentation}, and the post-fire digitisation of Notre-Dame de Paris~\cite{comte2024notredame}.
Packages such as Metashape~\cite{metashape}, RealityScan~\cite{realityscan}, 3DF~Zephyr~\cite{zephyr}, Pix4D~\cite{pix4d}, Meshroom~\cite{meshroom,openmvg}, and COLMAP~\cite{colmap,colmapmvs} provide graphical interfaces for structure-from-motion and multi-view stereo.
At their core, these pipelines map posed images to a dense 3D surface, typically through multi-view stereo, although the exact reconstruction back-end of the proprietary packages is not publicly disclosed.
They are most reliable on global shape: fine relief can be recovered, but usually only under strong assumptions on viewpoint coverage and surface appearance, and the highest-frequency detail tends to be smoothed out by the multi-view stereo and meshing.
A further limitation concerns texture: because each surface point samples colour directly from the input photographs, capture illumination, including shadows and directional shading, is baked into the texture map (\cref{fig:shading}).
The resulting model therefore carries the acquisition lighting, which is undesirable when texture is meant to document intrinsic object appearance.

\paragraph{Reflectance transformation imaging.}
Multi-light acquisition is already familiar in the heritage community through reflectance transformation imaging (RTI) and polynomial texture maps~\cite{malzbender2001ptm,mudge2006rti,earl2010rti}, surveyed in~\cite{pintus2019mlic,mansouri2026rti}.
In RTI, an object is photographed from a single fixed viewpoint while lit successively from many directions, typically with an LED dome (\cref{fig:dome}) or a hand-held flash~\cite{tetzlaff2016flash,nam2018practical}.
The result is a stack of images with a fitted per-pixel reflectance model, from which the surface can be interactively relit~\cite{righetto2024neural} and faint features such as inscriptions or tool marks enhanced.
In its standard form, however, it remains a single-viewpoint, image-space technique: it neither produces 3D information nor explicitly disentangles the surface normals from the reflectance, the very quantities a multi-view fusion would need to build a single 3D representation.

\paragraph{Combining multi-view and multi-light in cultural heritage.}
Photogrammetry and RTI thus capture complementary cues, yet are mostly acquired and analysed separately.
Early surveys already captured both photogrammetry and RTI of the same object~\cite{miles2013photogrammetry}, and later efforts registered RTI onto a photogrammetric 3D model for joint viewing~\cite{pamart2019rti,pamart2020taco}; this shows both the interest of the combination and its difficulty, such multimodal integration requiring careful registration and remaining a recognised open challenge rather than a routine operation.
Other lines combine the two cues only post-hoc: one blends high-frequency dome photometric-stereo normals onto a low-frequency photogrammetric or laser height field~\cite{macdonald2016accuracy}, while another bakes RTI normal maps onto a coarse photogrammetric mesh, leaving the relief in a texture rather than the geometry~\cite{morita2024combined}; both keep the modalities in separate pipelines, registered and merged afterwards.
The approach we follow sidesteps this registration problem: instead of fusing RTI with photogrammetry as two separate modalities, we feed the multi-light images directly into photometric stereo and integrate the resulting per-view normals across viewpoints into a single 3D surface.

\paragraph{Photometric stereo.}
Photometric stereo recovers, at every pixel, the surface normal together with the reflectance from images of a fixed viewpoint under varying illumination~\cite{woodham1980}.
Classical formulations assume known, calibrated lighting; over the last few years the field has progressed rapidly on two fronts that matter for heritage.
First, the acquisition setting has been relaxed from the fully controlled laboratory dome towards semi-controlled, in-the-field capture: self-calibrated near-light methods recover the lighting on the fly from a coarse geometric proxy~\cite{coupryssvm}, while universal deep methods operate under entirely unknown, spatially-varying illumination~\cite{unimsps,sdmunips,linounips}.
Second, these methods increasingly handle complex materials, estimating not only a diffuse reflectance but richer, physically-based reflectance (PBR)~\cite{burley2012disney,sdmunips}.
By estimating a normal at every pixel from the illumination variation, photometric stereo reaches a higher level of surface detail than single-shot methods, including on non-Lambertian materials, sparser viewpoints, and weakly textured surfaces where photogrammetry is inherently ambiguous~\cite{karami2021noncollaborative}.
These advances make photometric stereo applicable well beyond the controlled dome, where it has begun to be benchmarked on dedicated data~\cite{dulecha2020synthps}.

\paragraph{Multi-view photometric stereo.}
In computer vision, MVPS was classically cast as a joint optimisation problem, deforming a mesh so that it both renders consistently across views and agrees with the per-view photometric-stereo normals~\cite{park2013}. Later methods estimated reflectance, normals, and lighting together for more general materials~\cite{diligentmv}.
These formulations had to balance conflicting objectives, such as a depth term against a normal term, and tended to smooth away the high-frequency relief encoded by the normals~\cite{KayaKOFG22,KayaKOFG23}.
Neural volume rendering changed this by supervising a neural signed distance function through a volumetric rendering loss~\cite{neus,neus2}; neural MVPS methods then gained markedly in detail by encouraging this field to agree with photometric-stereo normals~\cite{psnerf,nplmvps}, without re-parametrising the photometric quantities, since normals-only integration already recovers most of the relief~\cite{cao2024supernormal}.
RNb-NeuS~\cite{rnbneus,rnbneus2} keeps a single objective while additionally admitting reflectance as an optional cue, by encoding the heterogeneous per-view normals and reflectance into a common, radiance-like quantity that directly supervises the signed distance function.

\paragraph{Reflectance as shading-free texture.}
Because photometric stereo separates reflectance from shading, the colour map applied to the reconstructed mesh is the intrinsic reflectance of the surface, free of the shadows and shading baked in by single-shot photogrammetric texturing~\cite{rnbneus2}.
Recovering such a relightable, intrinsic appearance jointly with geometry is the shared goal of inverse-rendering methods, whether from single-illumination~\cite{nvdiffrec,iron,neuralpbir} or multi-light photometric~\cite{pir} capture; here it comes for free as a by-product of the chain.
For heritage documentation, this yields an illumination-neutral texture that does not lock the object into the lighting conditions of the acquisition~\cite{laurent2025combining}.

\section{Proposed Pipeline in Meshroom}
\label{sec:system}

\begin{figure}[t!]
  \centering
  \includegraphics[width=\textwidth]{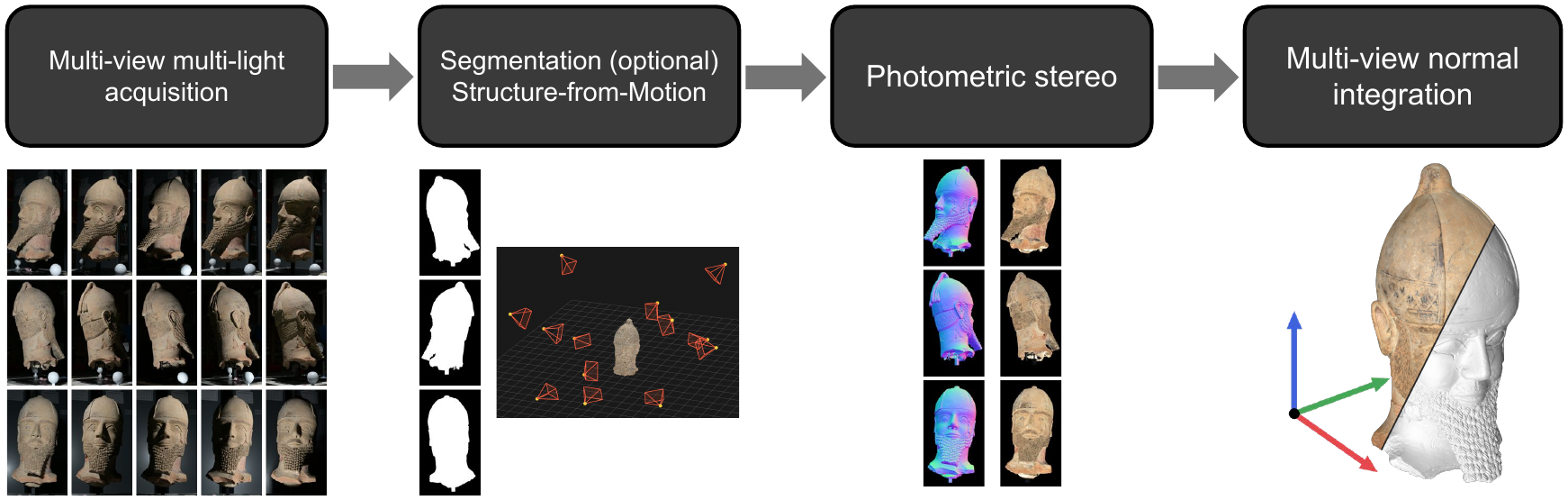}
  \caption{End-to-end data flow of the proposed pipeline. SAM~3 segments the multi-view, multi-light images, restricting the rest of the pipeline to the object of interest. Structure-from-motion recovers the camera parameters, optionally guided by the masks (\eg discard turntable-induced multiple motions), and an optional coarse geometry proxy that may be subsequently used by self-calibrated photometric stereo methods. Photometric stereo (calibrated, self-calibrated, or universal) turns the masked images into per-view normal maps, and optionally reflectance maps.
  Multi-view normal integration then fuses the per-view normals and reflectance with the camera parameters into a single mesh textured with the recovered reflectance.}
  \label{fig:overview}
  \vspace{-10pt}
\end{figure}

We now present the proposed pipeline: a multi-view, multi-light surface reconstruction chain assembled as nodes inside the open-source photogrammetry software Meshroom~\cite{meshroom}, turning illumination-varying image sets into a single textured, high-fidelity mesh.
\Cref{fig:overview} gives an overview of the data flow, from acquisition to the final mesh, and the rest of this section follows it stage by stage.

\subsection{Multi-view Multi-light Acquisition}
\label{sec:system:acquisition}

\begin{figure}[t]
  \centering
  \begin{tabular}{@{}c@{\hspace{8pt}}c@{}}
    \includegraphics[height=4.1cm]{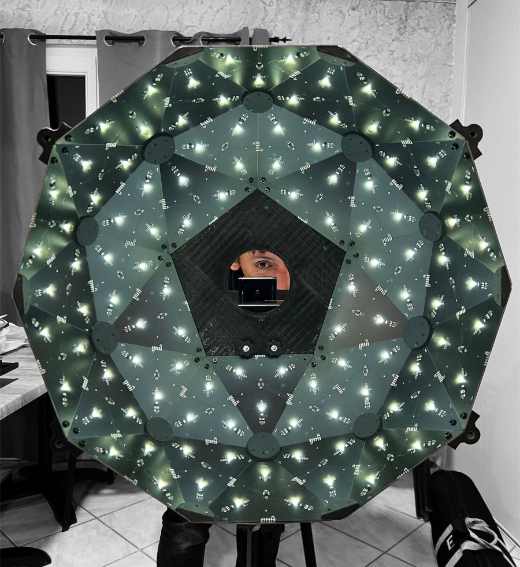} &
    \includegraphics[height=4.1cm]{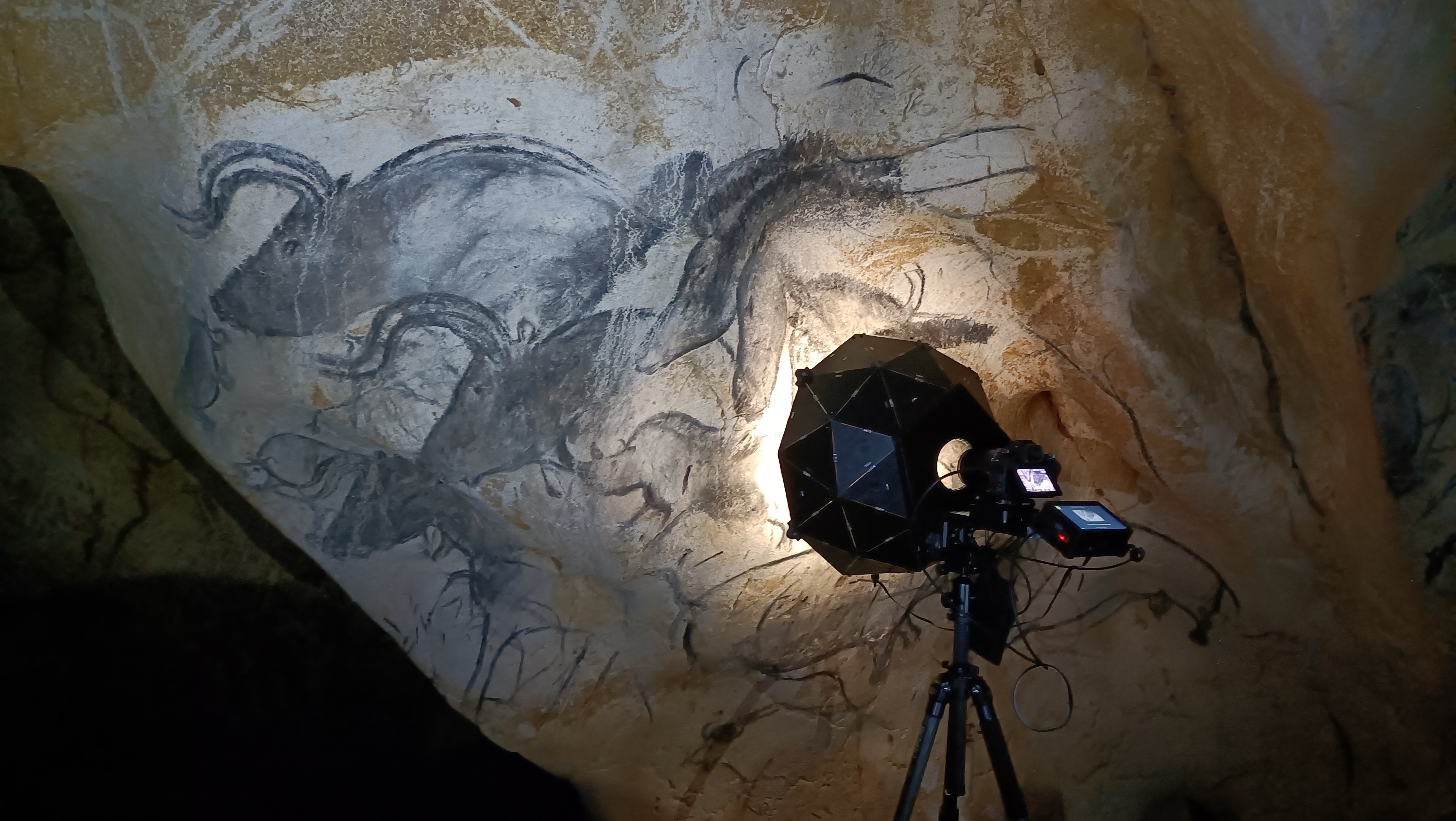} \\
  \end{tabular}
  \caption{Multi-light acquisition with an RTI dome (here a Mercurio Imaging dome~\cite{mercurio}). \emph{Left}: the dome seen from the front, its 105 LEDs distributed over a hemisphere around a central camera; firing the LEDs one at a time yields one image per light direction. \emph{Right}: such a dome in use on the \emph{Panneau des Chevaux} in the Chauvet cave.}
  \label{fig:dome}
  \vspace{-10pt}
\end{figure}

Multi-view, multi-light acquisition couples several viewpoints with several illuminations per viewpoint, the latter already familiar to the heritage community through RTI domes (\cref{fig:dome}) or a hand-held flash (\cref{sec:related}).
The pipeline accepts data organised per viewpoint and per light, captured with such a dome, an open-hardware camera-and-turntable setup, or a hand-held flash in the field, and places no constraint on image resolution (45.7\,MP in our experiments).

\subsection{A Nodal Pipeline in Meshroom}
\label{sec:system:pipeline}

We build on the nodal architecture of Meshroom~\cite{meshroom}: a reconstruction pipeline is a directed acyclic graph of nodes, each wrapping a tool, with an intermediate cache, automatic invalidation of downstream nodes, and standard file formats exchanged between nodes.
Meshroom (version~2025.1 and above) already reads the formats heritage acquisition produces, from camera RAW (\eg Canon CR2, Nikon NEF, Sony ARW) to high-dynamic-range OpenEXR through the OpenImageIO library, and can fuse bracketed exposures into HDR~\cite{debevec1997recovering}; it scales to the very-high-resolution datasets typical of photogrammetry and can dispatch its stages, including the per-pose photometric stereo computations, across a render farm through a pluggable submitter mechanism (\eg Tractor~\cite{tractor}).
We chose Meshroom for this extensibility and its mature, widely used interface: it lets us graft a complete photometric stereo and multi-view normal integration chain onto the existing structure-from-motion stage, making the methods accessible to end users while keeping the pipeline adaptable to each acquisition setup.
Every intermediate result, masks, camera parameters, per-view normal and reflectance maps, and the final mesh, can be visualised and reused.

Our contributions are the photometric stereo nodes, the multi-view normal integration node and the automatic masking, but also their orchestration with the existing components: feeding the structure-from-motion coarse proxy to self-calibrated photometric stereo, inserting light-calibration nodes where a method requires them, and deriving the metric scaling and centering of the neural volume from the object masks or the sparse point cloud.
We release all these nodes, together with several ready-to-use MVPS pipelines, as open-source Meshroom plugins implemented in Python: practitioners with no computer vision background can run the prebuilt pipelines as is, while more technical users can recompose the nodes, or implement new ones in Python, for their own acquisitions.

\paragraph{Automatic object masking.}
Heritage objects are usually imaged on a turntable, in a light tent, or on a stand for objects that need several faces to be captured, and a binary mask separating the object from the background serves two distinct purposes.
First, on a turntable the background and the object move differently from one shot to the next, a standard failure case for structure-from-motion; masking the background removes these conflicting motions.
Second, many neural surface reconstruction methods, including the one we rely on, take per-view object masks as input to bound the reconstructed volume~\cite{neus,rnbneus2}.
Producing such masks by hand, on every view, is tedious.
We therefore integrate automatic semantic segmentation as a Meshroom node, \texttt{ImageSegmentationSam3}~\cite{mrsegmentation}, built on the Segment Anything model with concepts (SAM~3)~\cite{carion2025sam3segmentconcepts}. From a lightweight prompt naming the object, optionally refined with positive and negative bounding boxes, it produces a per-view mask that then propagates through the whole chain (structure-from-motion, multi-view stereo, photometric stereo and integration), making object-centric reconstruction effectively automatic.

\paragraph{Structure-from-motion and multi-view stereo.}
These stages are already integrated into Meshroom through AliceVision~\cite{meshroom}, which has evolved substantially beyond the initial OpenMVG~\cite{openmvg} and CMPMVS~\cite{cmpmvs} implementations.
They detect and match image features across the input views, recover the camera parameters together with a sparse point cloud, and can optionally build a dense surface proxy.
Fiducial markers (CCTag~\cite{cctag} or AprilTag~\cite{apriltag}) can also be placed in the scene, for two purposes.
They improve robustness on objects that are hard for structure-from-motion: for a weakly textured piece on a turntable, markers on the turntable anchor the camera poses through reliable correspondences rather than uncertain matches on the object.
Given their known 3D coordinates, Meshroom also uses them to set the metric scale, via its \texttt{SfMTransform} node. %
The recovered camera parameters anchor the subsequent multi-view normal integration; the optional coarse surface proxy is used only by the self-calibrated near-light photometric stereo node~\cite{coupryssvm} for its light estimation.

\paragraph{The photometric stereo ecosystem.}
We implement the photometric stereo stage as interchangeable Meshroom nodes, covering calibrated lighting recovered from calibration spheres, self-calibrated near-light illumination, and unknown lighting handled by universal neural methods.
Exposing these alternatives behind a common interface lets practitioners select and compare the method that best matches the acquisition at hand.
All nodes output per-view normal maps, with reflectance maps provided when supported by the selected method.

\begin{figure}[t!]
  \centering
  \begin{tabular}{@{}>{\centering\arraybackslash}m{0.31\textwidth}@{\hspace{8pt}}>{\centering\arraybackslash}m{0.63\textwidth}@{}}
    \includegraphics[width=\linewidth]{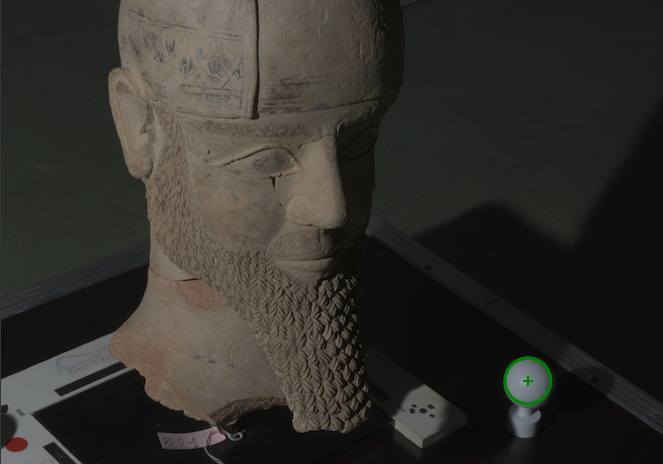} &
    \includegraphics[width=\linewidth]{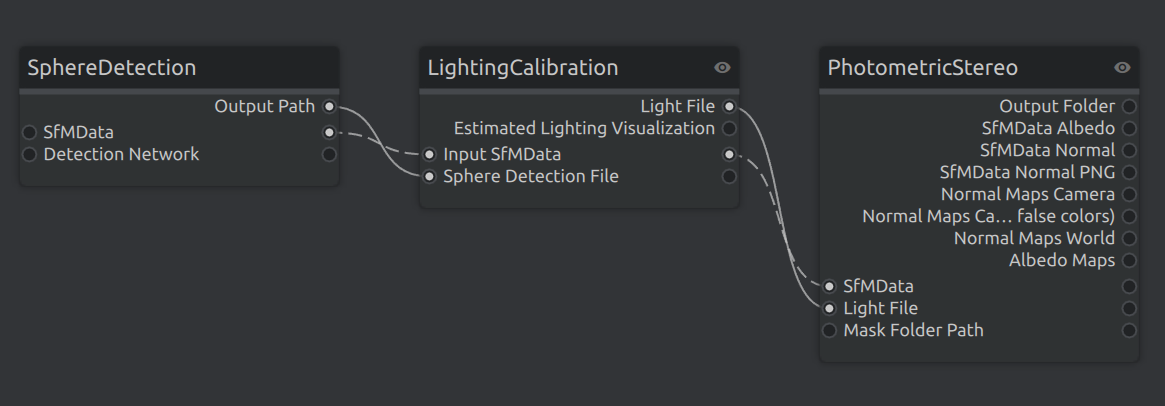} \\
    {\footnotesize (a) Automatic sphere detection} &
    {\footnotesize (b) Calibrated photometric-stereo sub-pipeline} \\
  \end{tabular}
  \caption{Calibrated photometric stereo in Meshroom. (a) A calibration sphere placed in the scene, here beside the Cypriot warrior head, is detected automatically (green circle) by the \texttt{SphereDetection} node. (b) The calibrated sub-pipeline, \texttt{SphereDetection}\,$\rightarrow$\,\texttt{LightingCalibration}\,$\rightarrow$\,\texttt{PhotometricStereo}, recovers the per-image light directions from the sphere and then the per-pixel normals and reflectance.}
  \label{fig:spheredet}
  \vspace{-10pt}
\end{figure}

Calibrated photometric stereo, the most classical and elementary setting, assumes the light directions and intensities are known.
In our pipeline, a calibration sphere of known radius is placed in the field of view (\cref{fig:spheredet}) and detected automatically (\texttt{SphereDetection}~\cite{fainsinspheres}); the \texttt{LightingCalibration} node locates the specular highlight on the sphere and infers each light direction (directional-light assumption in the current implementation), and the \texttt{PhotometricStereo} node then applies classical calibrated photometric stereo~\cite{woodham1980} to recover the per-pixel normals and reflectance.

Self-calibrated near-light photometric stereo~\cite{coupryssvm} removes this calibration step for nearby light sources: it recovers the position, intensity and anisotropy parameter of each light on the fly from a geometric proxy of the scene, either the sparse point cloud or, preferably, the coarse surface mesh of the multi-view stereo node~\cite{cmpmvs}, whose denser geometry yields more accurate surface normals for the light estimation.
This fits in-the-field acquisition, for example with a hand-held flash, where nothing about the lighting is measured.
As the state of the art in its category and based on an explicit physical model rather than learned from data, it introduces no learned-dataset bias and scales well, though it is best suited to near-Lambertian surfaces.

Unlike the calibrated and self-calibrated near-light approaches, universal photometric stereo requires neither measured lighting nor a coarse geometric proxy: it relies on learned neural models to recover surface normals, and reflectance when supported, under unknown, spatially varying illumination and complex, non-Lambertian materials.
We integrate three universal photometric-stereo methods that together represent the state of the art with complementary trade-offs, each wrapped as a Meshroom node: UniMS-PS~\cite{unimsps,mrunimsps} handles very-high-resolution inputs at multiple scales, at a higher computational cost; SDM-UniPS~\cite{sdmunips,mrsdmunips} additionally recovers a physically-based reflectance; and LINO-UniPS~\cite{linounips,mrlinounips}, the most recent, explicitly decouples illumination to preserve high-frequency detail, running in minutes where UniMS-PS takes hours on the same GPU.

Taken together, these nodes span the full range of lighting regimes, from calibrated to fully uncalibrated, and all feed the same multi-view normal integration.
Their value lies in this interoperability and in making research-prototype methods available, and directly comparable, to practitioners.
The calibrated node is kept for a reason: when the acquisition can be controlled, the more the lighting is constrained and known, the better-posed the reconstruction, which matters for metrology-oriented documentation.

\paragraph{Multi-view normal-and-reflectance integration.}
The final node fuses the per-view normal and reflectance maps into a single surface.
Following recent neural surface reconstruction, it represents the scene as a signed distance function (SDF) whose zero level set is the surface, and optimises it so that the SDF gradient agrees, in each view, with the per-view normal maps from photometric stereo~\cite{rnbneus2, cao2024supernormal}; the reflectance is folded into the same objective by re-parametrising normals and reflectance as a simulated radiance field.
The mesh is then extracted by marching cubes~\cite{lorensen1987marching}.
We integrate RNb-NeuS2~\cite{rnbneus2,mrrnbneus2}, which adapts the instant neural surface reconstruction of NeuS2~\cite{neus2} to normal-and-reflectance inputs: its single objective includes the normals-only integration of SuperNormal~\cite{cao2024supernormal} as a special case (see~\cite{rnbneus2}, Sec.~5.3), runnable as an option, while additionally integrating the reflectance, for instance from SDM-UniPS, that a normals-only integrator would discard.
On the implementation side, its multiresolution hash encoding~\cite{instantngp} and CUDA backend cut the per-scene optimisation from about fifteen hours in the original RNb-NeuS~\cite{rnbneus} to roughly five minutes.
The result is a single high-fidelity mesh textured with the reflectance, free of baked-in shading.
RNb-NeuS2 carries an NVIDIA license restricting it to academic use; for commercial settings, we contribute OpenRNb~\cite{openrnb} (MPL~2.0), a from-scratch reimplementation of the integration backend exposed as a drop-in alternative.

\section{Results on Real Archaeological Objects}
\label{sec:results}

\begin{figure}[htb]
  \centering
  \setlength{\tabcolsep}{2pt}
\renewcommand{\arraystretch}{0.9}
\begin{tabular}{@{}*{5}{>{\centering\arraybackslash}m{0.19\linewidth}}@{}}
  \includegraphics[width=\linewidth]{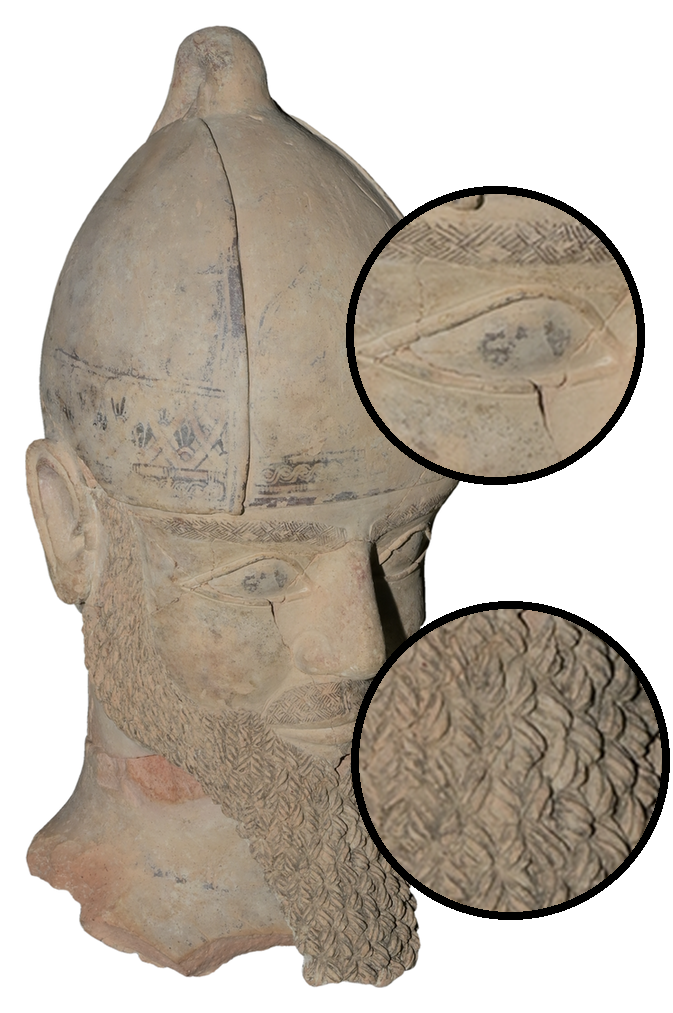} &
  \includegraphics[width=\linewidth]{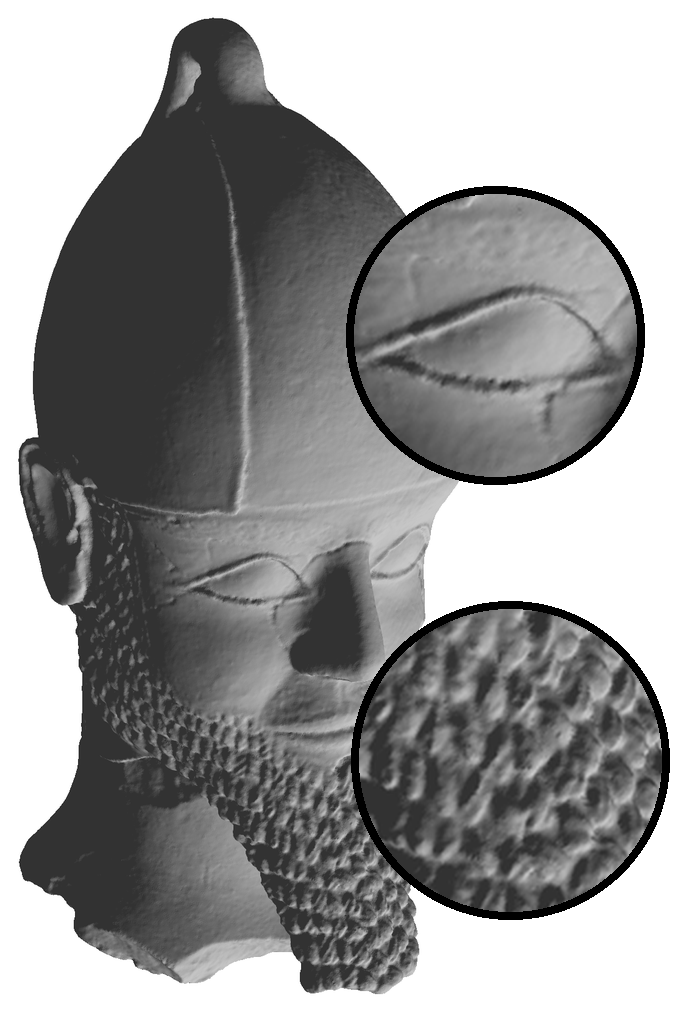} &
  \includegraphics[width=\linewidth]{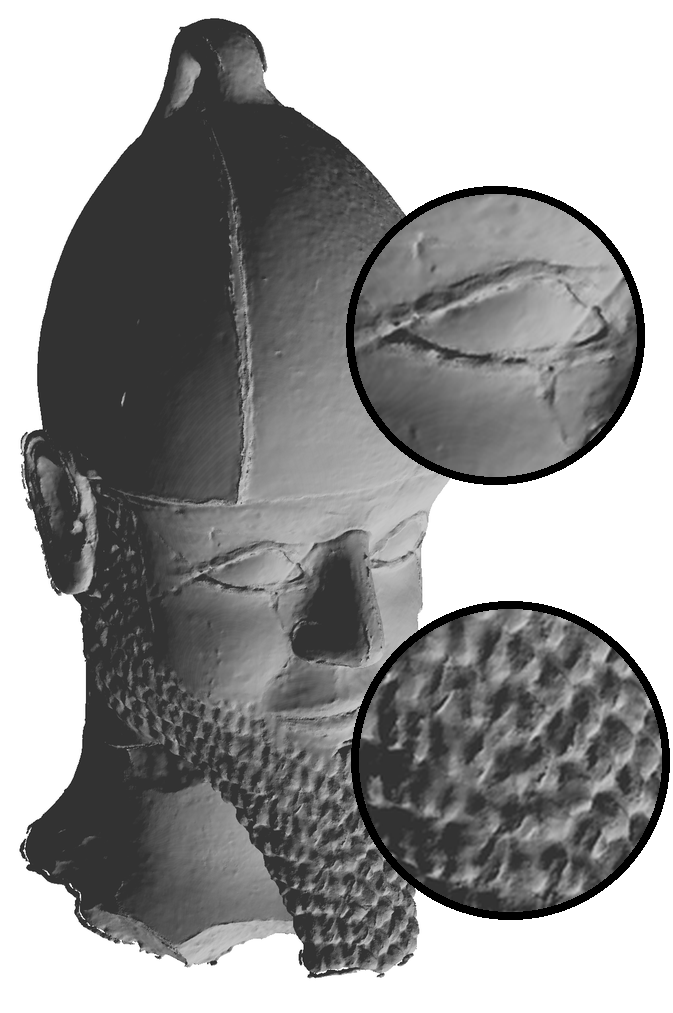} &
  \includegraphics[width=\linewidth]{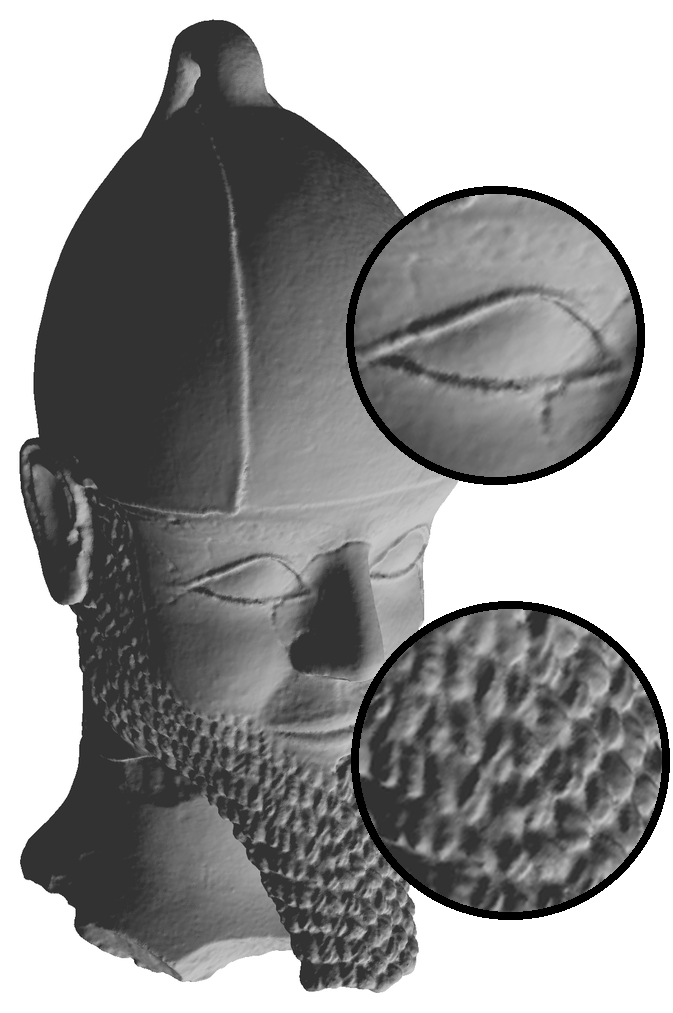} &
  \includegraphics[width=\linewidth]{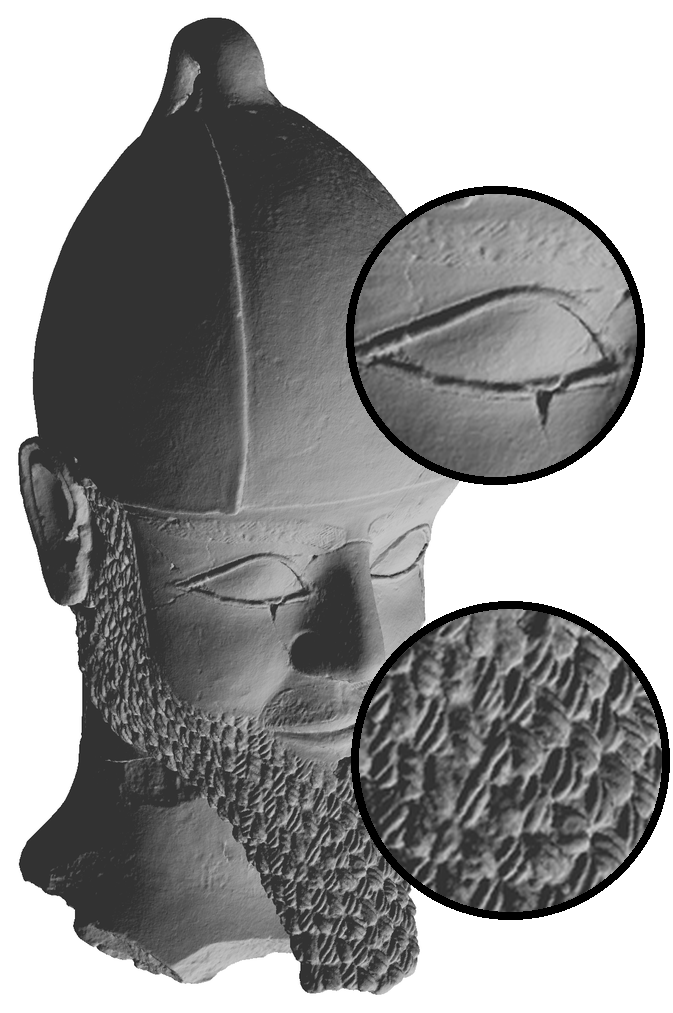} \\
  {\footnotesize Reference} &
  {\footnotesize Metashape} &
  {\footnotesize COLMAP} &
  {\footnotesize Meshroom} &
  {\footnotesize Proposed}\\ %
\end{tabular}

  \caption{Head of a Cypriot warrior (terracotta, 6th century BC), under a matched viewpoint. From left to right: a reference photograph, three photogrammetric reconstructions from 121 views, namely Metashape~\cite{metashape}, COLMAP~\cite{colmap,colmapmvs} and the Meshroom photogrammetry pipeline~\cite{meshroom}, and the proposed pipeline with self-calibrated near-light PS~\cite{coupryssvm}, integrated by RNb-NeuS2~\cite{rnbneus2} (13 views under 12 illuminations each, namely 156 images). The black insets magnify the eye (top) and the braided beard (bottom). The reference is consistent with the recovered relief, supporting that it is genuine rather than hallucinated; the proposed pipeline restores sharper relief, notably the separated strands of the beard, at a comparable image budget.
  }
  \label{fig:warrior}
  \vspace{-10pt}
\end{figure}

We evaluate the pipeline qualitatively on four real archaeological objects, chosen so that each exercises a different instantiation of the pipeline, namely a different photometric stereo method.\footnote{The head of a Cypriot warrior and the woman with a calathos are held by the Mus\'ee Saint-Raymond, Toulouse; the engraved bone comes from the Gourdan cave.}
In these experiments, all multi-view photometric stereo results are produced with the RNb-NeuS2 integration node~\cite{rnbneus2} (\cref{sec:system}).
All datasets were captured with a full-frame 45.7\,MP camera, and the full chain, raw development, structure-from-motion, photometric stereo and integration, runs in a few hours per object on a single GPU, comparable to a conventional photogrammetric reconstruction.
Because fine engravings and tool marks are often central to documentation, we focus on the recovered surface relief, rendering the reconstructed geometry under controlled shading rather than its colour texture.
A structured-light scan~\cite{artec,artecmicro}, when available, is included as an additional comparison modality and never as a target: it is one digitisation technique among others, relying on dedicated hardware rather than the accessible, image-based setting we address.

\subsection{Head of a Cypriot Warrior}
\label{sec:results:warrior}

The first object is a terracotta head whose fine, high-frequency relief, concentrated in the hair and beard, is the kind of structure known to be hard to reconstruct faithfully.
We compare the proposed pipeline against photogrammetry at a comparable image budget, rendering the geometry alone (\cref{fig:warrior}), using self-calibrated near-light photometric stereo~\cite{coupryssvm} for the per-view normals, with a 35\,mm lens and a hand-held flash for the multi-light captures.
For this almost identical number of images, the proposed pipeline recovers markedly finer relief: as the eye and beard insets show, it separates the strands of the braided beard that dense matching tends to fuse and smooth.
The reference photograph, shown alongside, confirms that this relief is genuinely present on the object and not a neural hallucination.

\subsection{Engraved Bone of Gourdan}
\label{sec:results:bone}

\begin{figure}[t]
  \centering
  \setlength{\tabcolsep}{2pt}
\renewcommand{\arraystretch}{1.0}
\begin{tabular}{@{}*{2}{>{\centering\arraybackslash}m{0.49\linewidth}}@{}}
  \includegraphics[width=\linewidth]{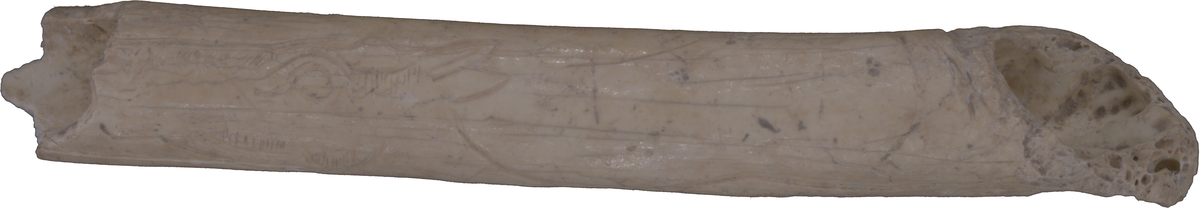} &
  \includegraphics[width=\linewidth]{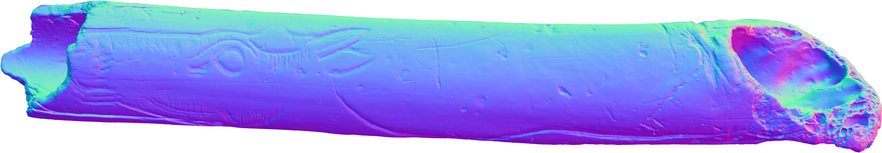} \\
  {\footnotesize Reference} &
  {\footnotesize Normal map (PS~\cite{unimsps})} \\[2pt]
  \includegraphics[width=\linewidth]{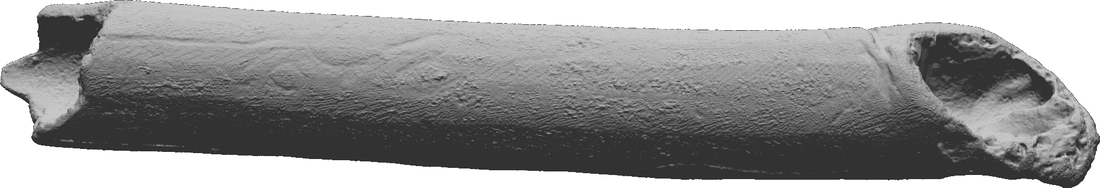} &
  \includegraphics[width=\linewidth]{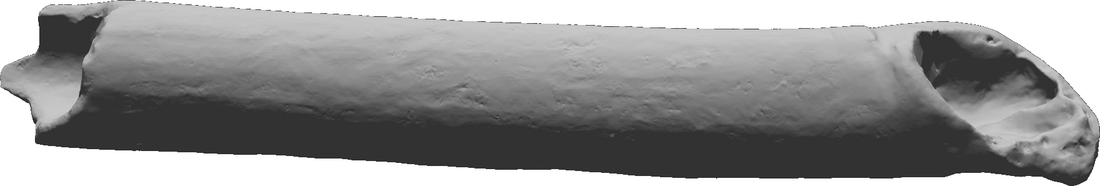} \\
  {\footnotesize Metashape} &
  {\footnotesize RealityScan} \\[2pt]
  \includegraphics[width=\linewidth]{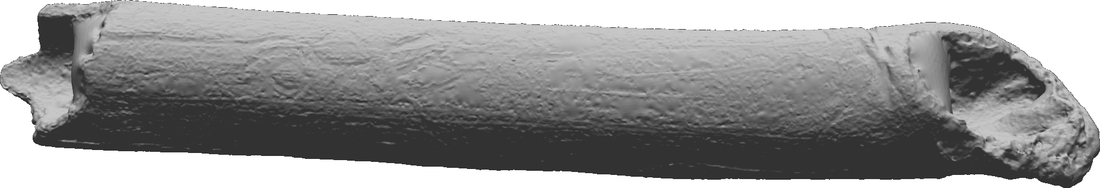} &
  \includegraphics[width=\linewidth]{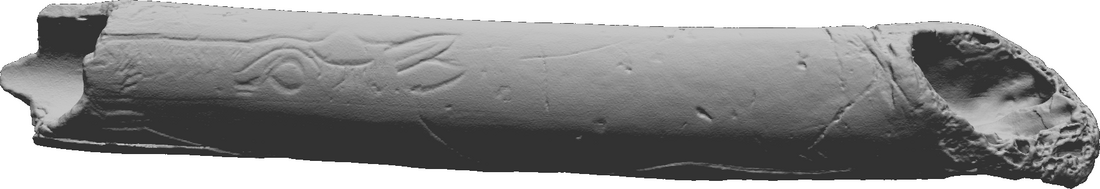} \\
  {\footnotesize COLMAP} &
  {\footnotesize Proposed}\\ %
\end{tabular}

  \caption{Engraved bone from the Gourdan cave, rendered under grazing light: a reference photograph, the UniMS-PS~\cite{unimsps} per-view normals, three photogrammetric reconstructions (RealityScan~\cite{realityscan}, Metashape~\cite{metashape} and COLMAP~\cite{colmap,colmapmvs}), and the surface integrated from these normals by RNb-NeuS2~\cite{rnbneus2}. The engravings are barely visible with photogrammetry but recovered by the proposed pipeline.
  }
  \label{fig:bone}
\end{figure}

The second object is an engraved bone whose surface bears shallow incised engravings, the kind of faint, high-frequency relief that a flat photograph or a smoothed mesh would lose.
It is acquired from 23 viewpoints under 35 illuminations each, with a 50\,mm lens and an RTI dome~\cite{mercurio}, and reconstructed without any photogrammetric prior: the per-view normals are estimated by the universal method UniMS-PS~\cite{unimsps} and integrated into a single mesh by RNb-NeuS2~\cite{rnbneus2}.
The recovered normal map already makes the engravings legible, and the integrated surface preserves them in 3D (\cref{fig:bone}), showing that the universal photometric stereo nodes make the pipeline able to digitise surfaces without any light calibration.
By contrast, photogrammetric reconstructions from the 23 views alone (RealityScan~\cite{realityscan}, Metashape~\cite{metashape} or COLMAP~\cite{colmap,colmapmvs}) leave the engravings barely visible, confirming that single-light multi-view stereo does not resolve this faint relief.

\begin{figure}[t]
  \centering
  \setlength{\tabcolsep}{2pt}
\renewcommand{\arraystretch}{0.9}
\begin{tabular}{@{}*{4}{>{\centering\arraybackslash}m{0.245\linewidth}}@{}}
  \includegraphics[width=\linewidth]{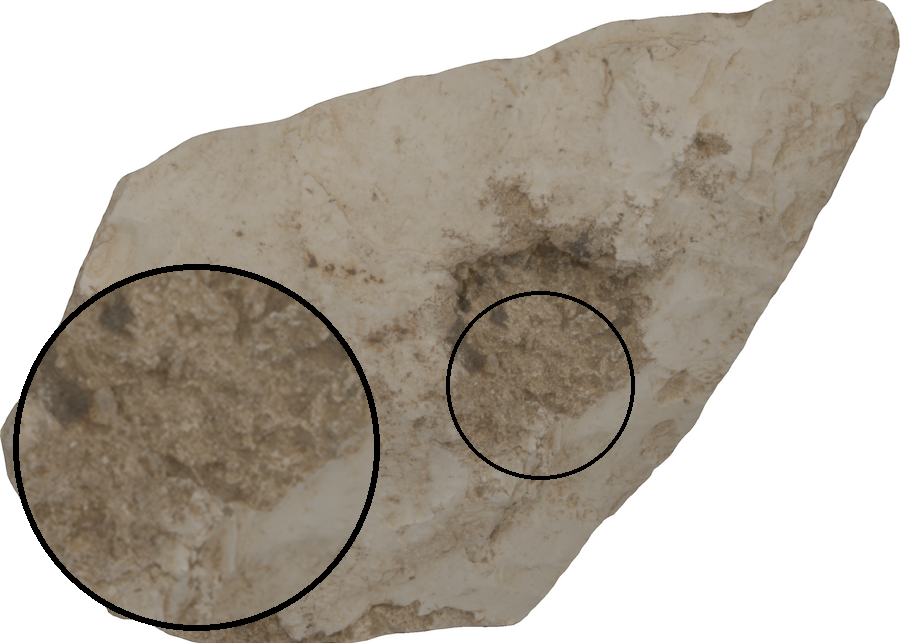} &
  \includegraphics[width=\linewidth]{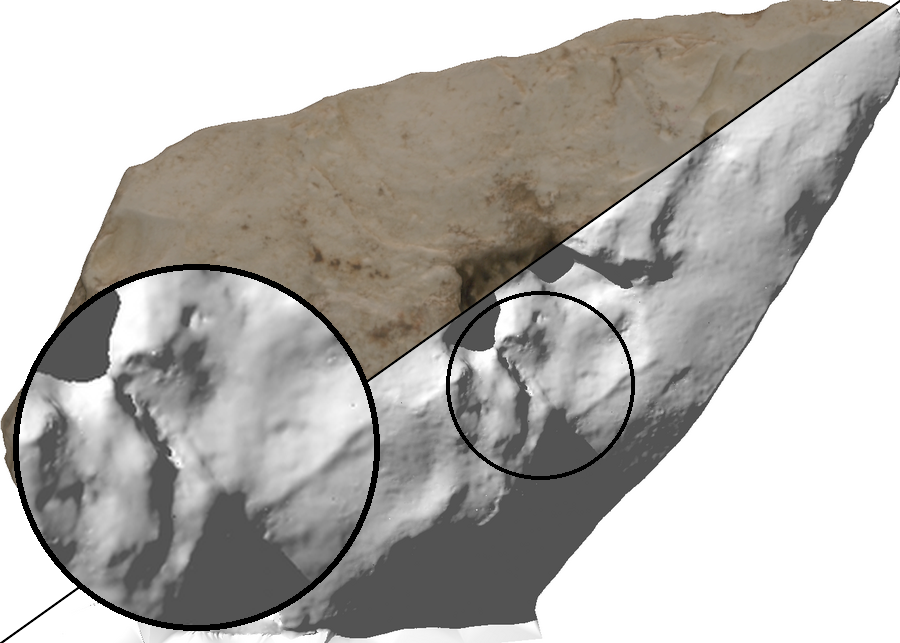} &
  \includegraphics[width=\linewidth]{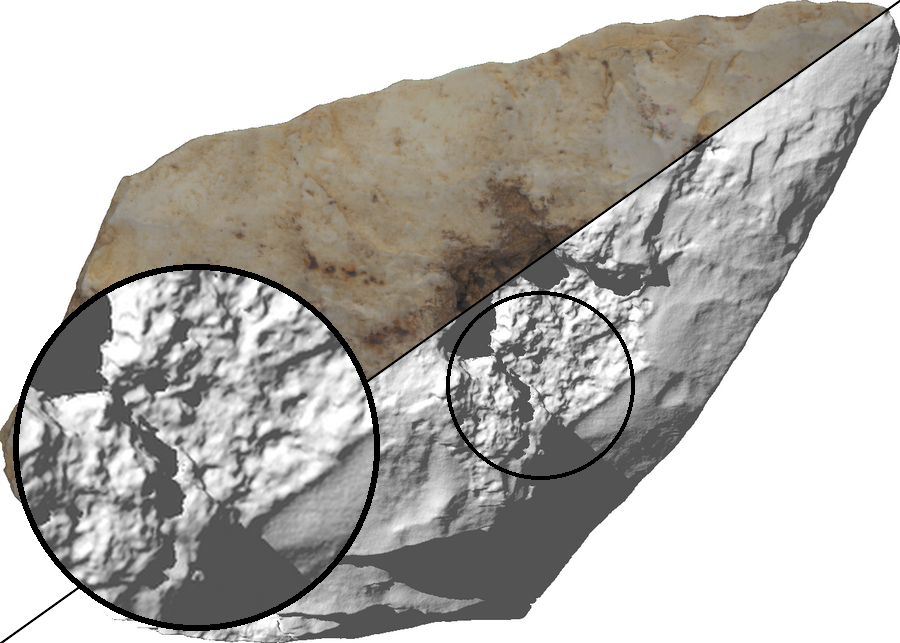} &
  \includegraphics[width=\linewidth]{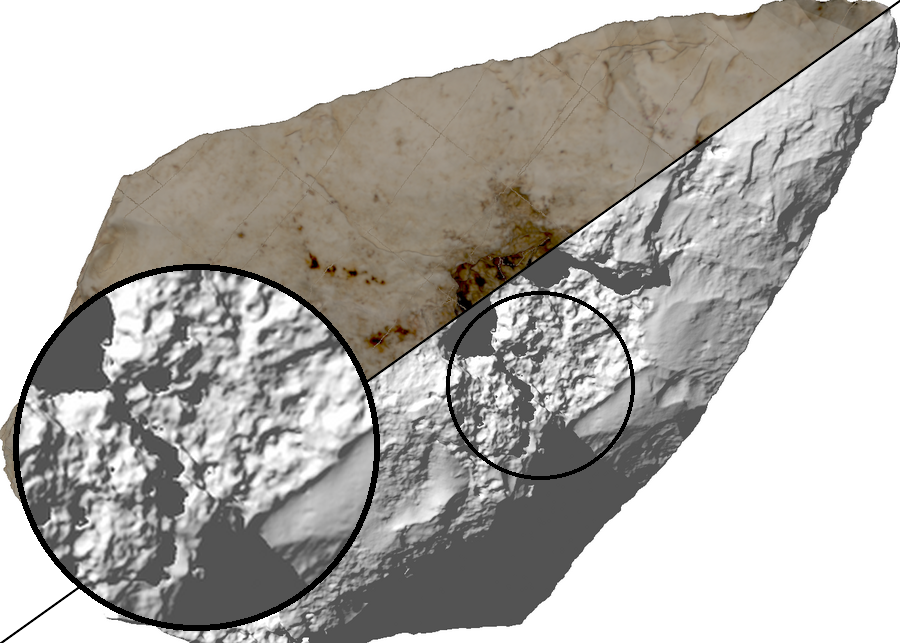} \\
  {\footnotesize Reference} &
  {\footnotesize RealityScan} &
  {\footnotesize Proposed} & %
  {\footnotesize Artec Micro~2} \\
\end{tabular}

  \caption{Lithic biface, from left to right: a reference photograph, RealityScan~\cite{realityscan}, the proposed pipeline, and an Artec Micro~2 scan~\cite{artecmicro}. For each method, a diagonal splits the relit texture (upper left) from the geometry alone (lower right); the circular inset magnifies the fine flake-scar relief of the boxed region.
  }
  \label{fig:biface}
  \vspace{-10pt}
\end{figure}

\subsection{Lithic Biface}
\label{sec:results:biface}

The third object is a flint biface whose surface is covered by shallow, intersecting flake scars: a low-amplitude, weakly textured relief that is hard to recover.
It also appears in \cref{fig:teaser}, reconstructed inside the Meshroom interface.
We compare the reconstructions in both geometry and textured appearance, relit under the illumination of a reference photograph (\cref{fig:biface}): each method panel is split along its diagonal, with the relit texture in the upper-left and the shading-only geometry in the lower-right.
The multi-light set comprises 12 viewpoints under 12 illuminations each (50\,mm lens, hand-held flash), against 162 photogrammetric views.
The proposed pipeline, self-calibrated near-light photometric stereo~\cite{coupryssvm} integrated by RNb-NeuS2~\cite{rnbneus2}, recovers the sharp ridges of the shallow flake scars and reproduces the textured appearance of the real object, at a level of detail very close to the structured-light scan (Artec Micro~2)~\cite{artecmicro}.

\subsection{Woman with a Calathos}
\label{sec:results:calathos}

\begin{figure}[htb]
  \centering
  \setlength{\tabcolsep}{2pt}
\renewcommand{\arraystretch}{0.9}
\begin{tabular}{@{}>{\centering\arraybackslash}m{0.035\linewidth}@{\hspace{1pt}}*{4}{>{\centering\arraybackslash}m{0.224\linewidth}}@{}}
  &
  {\footnotesize Reference} &
  {\footnotesize RealityScan} &
  {\footnotesize Proposed} & %
  {\footnotesize Artec Spider} \\
  \rotatebox{90}{\footnotesize Geometry} &
  \includegraphics[width=\linewidth]{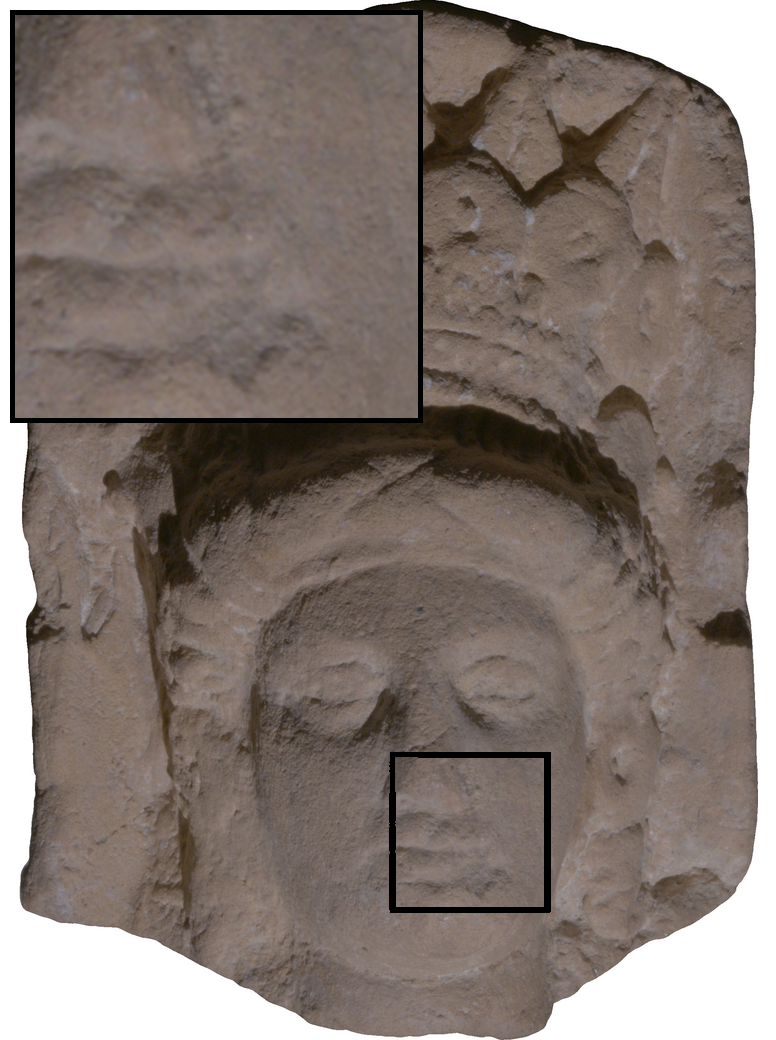} &
  \includegraphics[width=\linewidth]{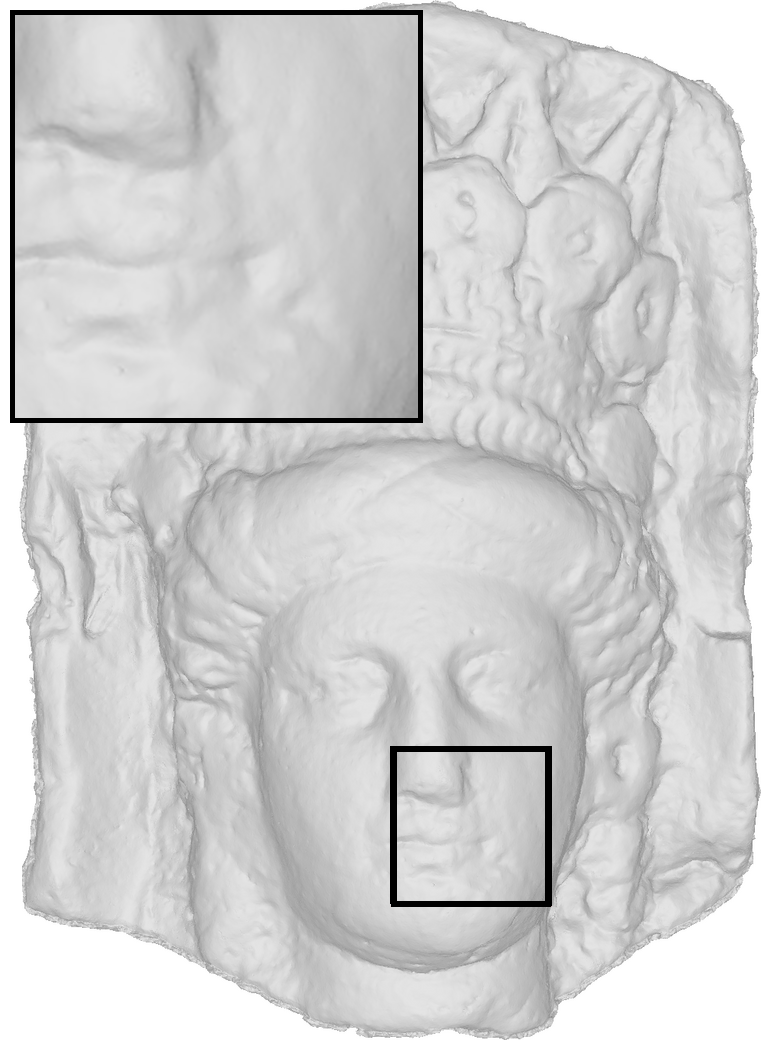} &
  \includegraphics[width=\linewidth]{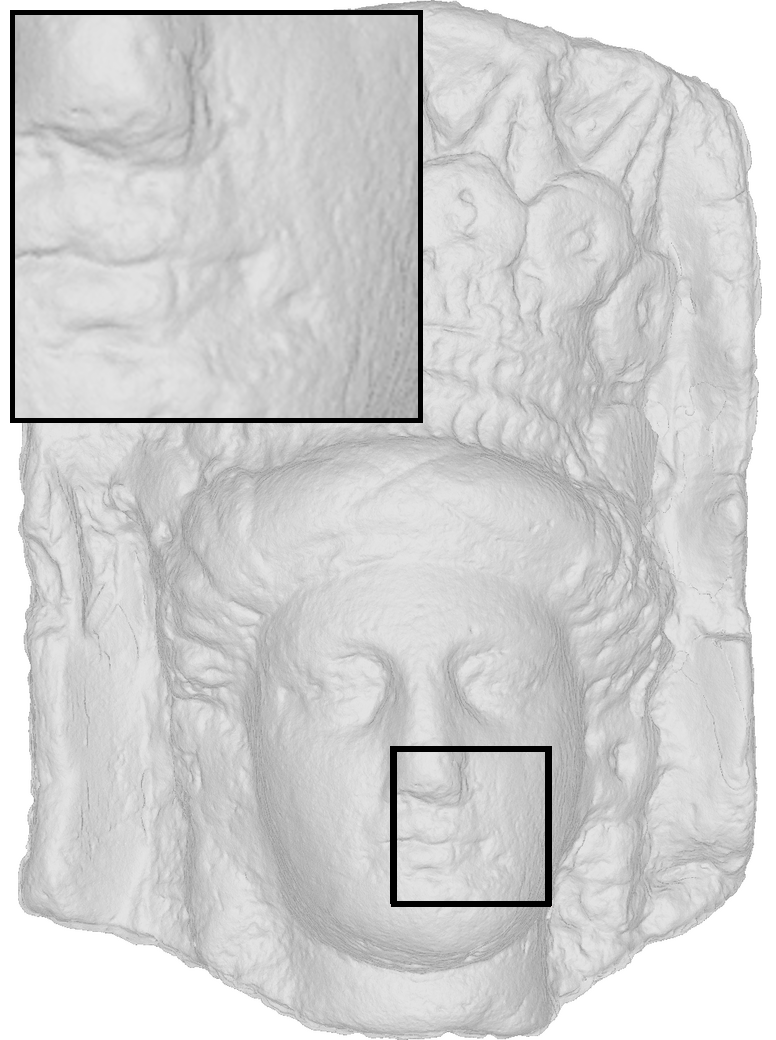} &
  \includegraphics[width=\linewidth]{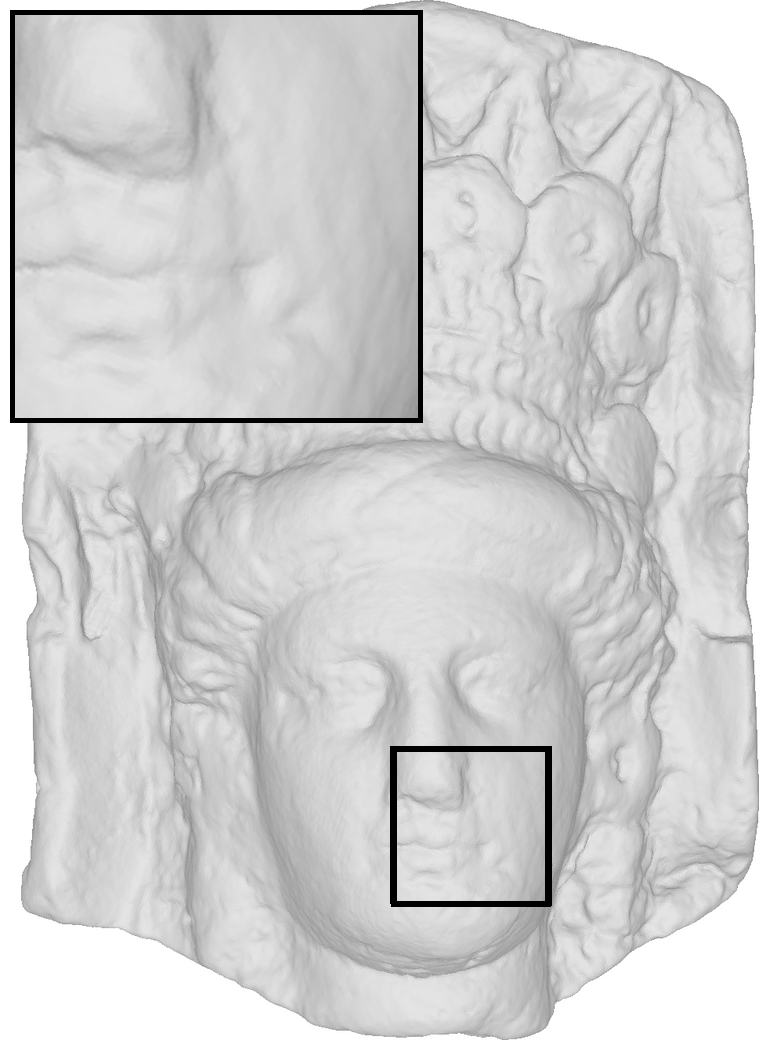} \\[1pt]
  \rotatebox{90}{\footnotesize Relit (grazing)} &
  \includegraphics[width=\linewidth]{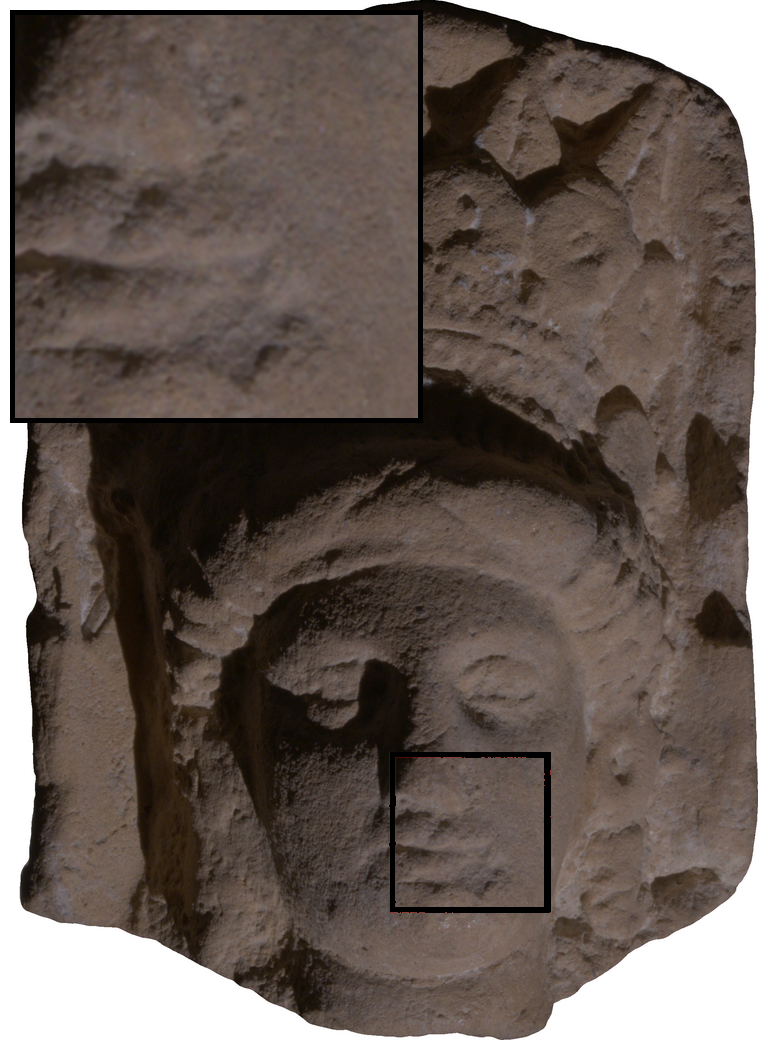} &
  \includegraphics[width=\linewidth]{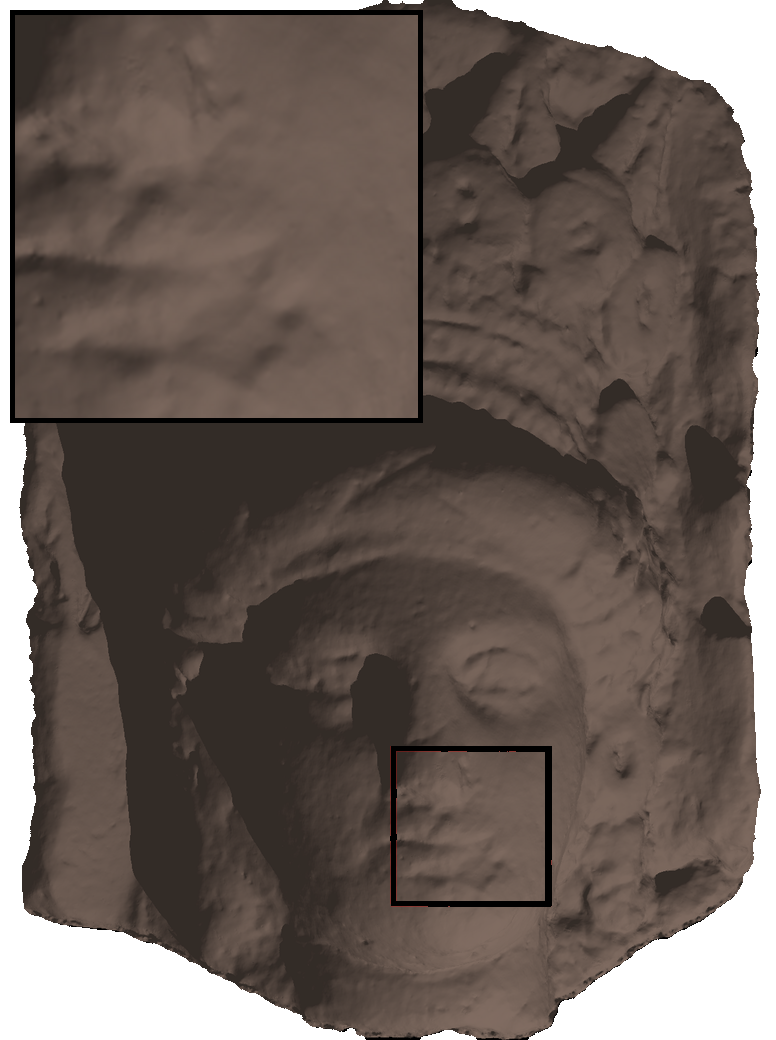} &
  \includegraphics[width=\linewidth]{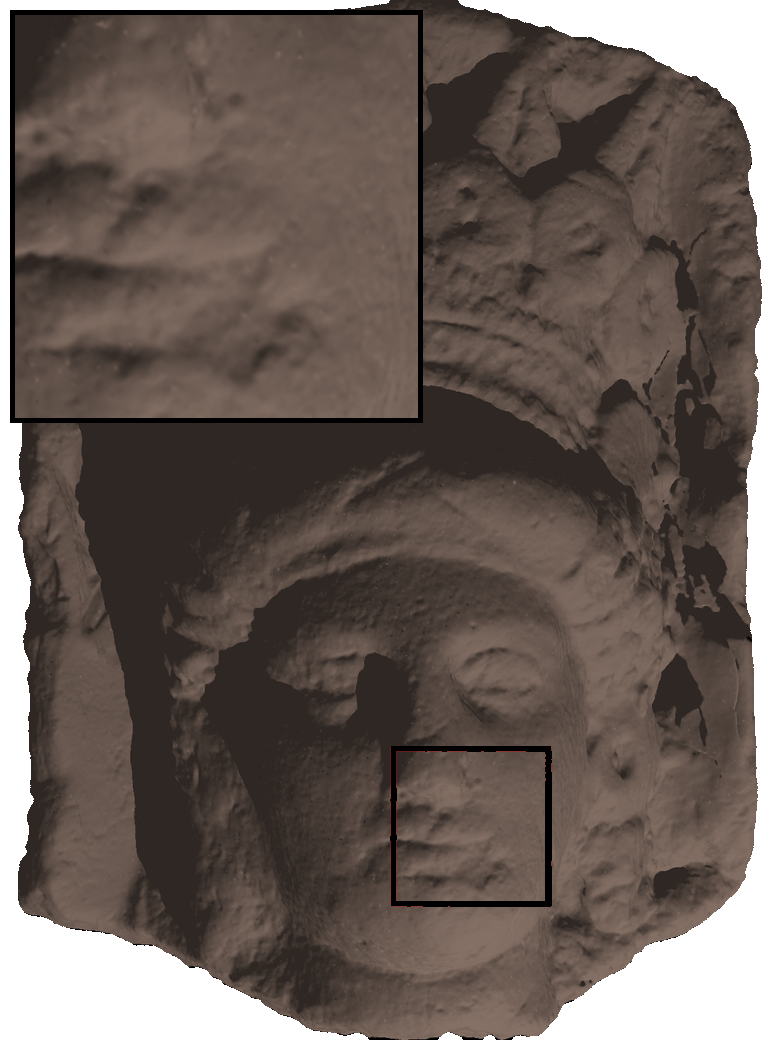} &
  \includegraphics[width=\linewidth]{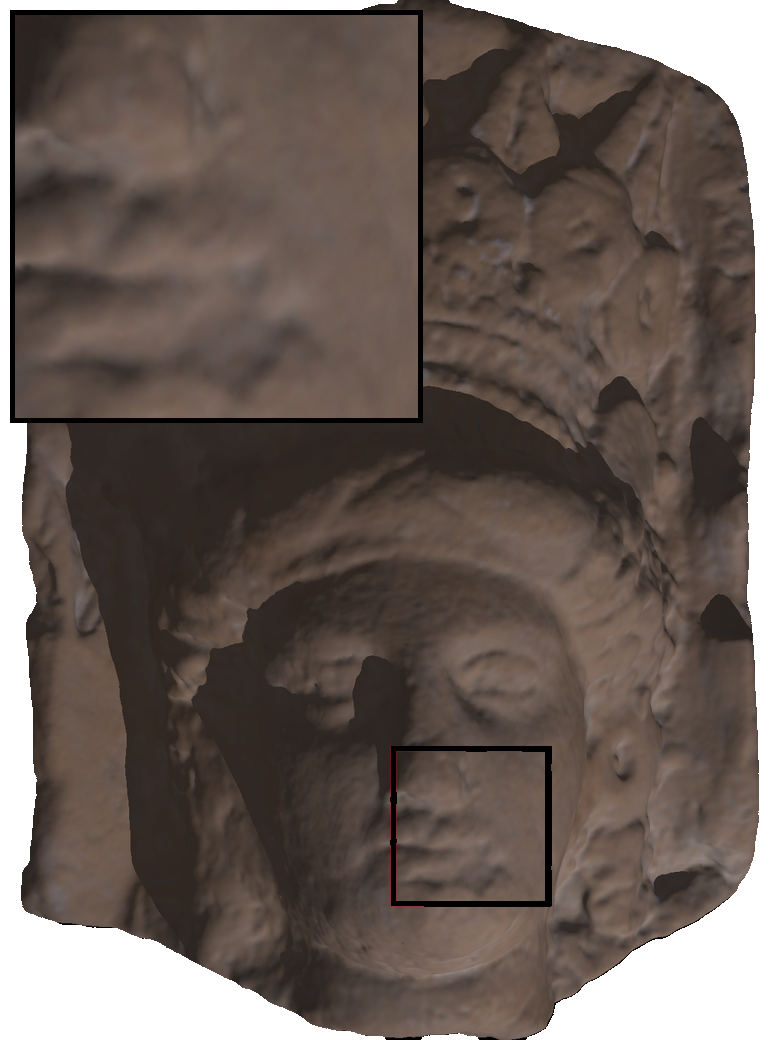} \\
\end{tabular}

  \caption{Woman with a calathos (ceramic, 4th century BC), with columns in correspondence. \emph{Top}: shading renders of the reconstructed geometry, next to a reference photograph. \emph{Bottom}: the same surfaces relit under a single grazing light matched to the reference, so that fine relief is revealed by the cast shadows; the black box magnifies the right corner of the mouth (inset, top left of each panel). Columns: reference, RealityScan~\cite{realityscan}, the proposed pipeline (PS~\cite{coupryssvm}, 8 views under 35 illuminations each) and a structured-light scan~\cite{artec}; photogrammetry uses 48 views. Among the shown modalities, the proposed pipeline best brings out the fine granularity of the ceramic micro-relief, including next to the structured-light scan at this resolution, and the grazing relighting matches the relief seen on the reference.
  }
  \label{fig:calathos}
  \vspace{-10pt}
\end{figure}

Finally, we turn to a ceramic head whose smooth, fine-grained surface carries a low-amplitude micro-relief that is easily lost.
We reconstruct it and show it next to a reference photograph (\cref{fig:calathos}): a photogrammetry pipeline (RealityScan~\cite{realityscan}), a structured-light scan (Artec Spider, with 3D point accuracy down to $50\,\mu$m)~\cite{artec}, and the proposed pipeline.
Photogrammetry uses 48 views, and the proposed pipeline uses 8 viewpoints, each imaged under 35 illuminations with a 50\,mm lens and an RTI dome~\cite{mercurio}, self-calibrated near-light photometric stereo~\cite{coupryssvm} providing the per-view normals and reflectance.
Rendered under identical shading (\cref{fig:calathos}, top), the proposed pipeline brings out the fine granularity of the ceramic surface that the other reconstructions, including the structured-light scan, do not render at this scale.
To check that this relief is genuine and not an artefact, we relight each surface under a single grazing light matched to the reference photograph (\cref{fig:calathos}, bottom): the micro-relief recovered by the proposed pipeline closely tracks the shadows cast on the real object, whereas the photogrammetric and scanned surfaces remain comparatively smooth.

\section{Discussion and Limitations}
\label{sec:discussion}

\paragraph{Limitations.}
Our integration still has limitations.
First, only a subset of the relevant methods is wrapped so far: further photometric stereo and integration methods, and emerging approaches that reconstruct the surface directly from the multi-light images without an explicit photometric-stereo-and-fusion split, are not yet available as nodes.
Second, several stages, the universal photometric stereo networks and parts of the multi-view stereo back-end, require an NVIDIA/CUDA GPU, which remains a deployment constraint for some heritage users.
Third, the multi-view integration node comes in two implementations that are not yet on par. RNb-NeuS2~\cite{rnbneus2} is the reference one but, building on NeuS2, it inherits a non-commercial NVIDIA research license; OpenRNb~\cite{openrnb} is a from-scratch, permissively licensed (MPL-2.0) reimplementation built on the open instant-nsr-pl framework~\cite{instantnsrpl}, and is therefore usable in commercial settings. On the five DiLiGenT-MV objects the two recover essentially the same sub-millimetre surfaces, yet OpenRNb's mean Chamfer distance is currently higher ($0.24\,mm$ against $0.18$\,mm on UniMS-PS normals, and $0.30\,mm$ against $0.22$\,mm on SDM-UniPS). This is promising for a fully open, commercially usable alternative, but closing this gap needs further investigation.
Finally, our evaluation on real archaeological objects is qualitative. Unlike the controlled DiLiGenT-MV benchmark reported above, these unique heritage objects have no metric ground truth, and the structured-light scans shown alongside are an alternative digitisation modality rather than a reference against which to compute an error. Quantifying reconstruction errors on such objects is one of the natural next steps to consolidate these findings.

\paragraph{Outlook.}
Several directions follow naturally.
The most immediate is to broaden the ecosystem, wrapping further photometric stereo and multi-view normal integration methods behind the same interface so that practitioners can compare them on their own data.
A more ambitious direction is to propagate a per-pixel reliability of the photometric estimates through the pipeline and expose them as a confidence map in the mesh texture, so that downstream analysis can flag regions to interpret with caution.
Because every intermediate output is exposed in open formats, the pipeline is also a natural basis for an open evaluation platform and for shared, frequency-resolved quality metrics.
Finally, since the system is embedded in Meshroom rather than introduced as a separate application, it inherits an interface and execution model already familiar from photogrammetric practice. A complementary user study with conservators and archaeologists would help refine presets, documentation, and acquisition guidelines for different heritage settings.

\section{Conclusion}
\label{sec:conclusion}

We have presented an end-to-end multi-view, multi-light pipeline assembled inside Meshroom, a widely used open-source photogrammetry framework.
Our contribution is the integration that turns a high-fidelity but lab-bound approach into an operational tool: a complete photometric-stereo ecosystem, calibrated, self-calibrated near-light and universal, automatic object masking, and a multi-view normal-and-reflectance integration node.
On real archaeological objects, the pipeline recovers fine surface relief and produces a shading-free texture, while reusing the interface and open formats that heritage practitioners already use for photogrammetry.
The system lowers the barrier to detailed 3D reconstruction within a tool already adopted in heritage. We hope this encourages wider use of multi-view photometric stereo, and its extension to a broader range of surface materials.

\section*{Acknowledgements}
Baptiste Brument was funded by the OPEN-DOPAMIn project (CNRS Innovation). This work was granted access to the Juliet computing platform, supported by a French government grant managed by the Agence Nationale de la Recherche under the ``Investissements d'avenir'' program (reference ``ANR-21-ESRE-0051'').
It has been supported by the University of Zurich and University Hospital Balgrist.

\bibliographystyle{splncs04}
\bibliography{main}

\end{document}